\documentclass[a4paper,11pt]{article}
\usepackage[english]{babel}
\usepackage{inputenc}
\usepackage[T1]{fontenc}
\usepackage{mathrsfs}
\usepackage[T1]{fontenc}
\usepackage{comment}
\usepackage{balance}
\usepackage{dsfont}
\usepackage[mono=false]{libertine}  
\usepackage[top=2.5cm, bottom=2.5cm, left=2.5cm, right=2.5cm]{geometry}
\usepackage{xcolor}

\usepackage{enumerate}
\usepackage{amsmath,amsthm,amssymb,amsfonts,bbm,bm,nicefrac,mathtools}
\usepackage{hyperref}

\usepackage[normalem]{ulem}
\definecolor{ps}{RGB}{0,0,200}

\usepackage{amsmath,amsfonts,bm}

\def\1{\bm{1}}

\DeclareMathAlphabet{\mathsfit}{\encodingdefault}{\sfdefault}{m}{sl}
\SetMathAlphabet{\mathsfit}{bold}{\encodingdefault}{\sfdefault}{bx}{n}

\newcommand{\E}{\mathbb{E}}

\newcommand{\proba}{\mathbb{P}}

\newtheorem{theorem}{Theorem}

\usepackage[ruled,vlined]{algorithm2e}
\usepackage{url}
\usepackage{algpseudocode}

\title{On the Slow Convergence to Trivial Solutions of Algorithms \\ for Hard Optimization Problems}

\author{
    Ali Hussaini Umar$^{1}$\thanks{Corresponding author: \texttt{aumar@sissa.it}}
    \quad
    Jean Barbier$^{2}$
    \quad
    Matthieu Jonckheere$^{3}$
    \quad
    Manuel S\'aenz$^{4}$
    \\[0.5em]
    \small $^{1}$SISSA, Trieste, Italy. 
    \small $^{2}$International Center for Theoretical Physics, Trieste, Italy.\\    
    \quad
    \small $^{3}$LAAS--CNRS, Toulouse, France.
    \small $^{4}$Universidad de San Andr\'es, Argentina.
}

\begin{document}

\maketitle

\begin{abstract}
    Hard combinatorial optimization problems, many of which are NP-hard, present fundamental algorithmic challenges. Average-case analysis on random instances has emerged as a powerful framework for understanding typical algorithmic performance beyond worst-case guarantees. A substantial body of work has established negative results: for sufficiently hard instances (often controlled by the underlying graph connectivity/constraints density), no known polynomial-time algorithm can significantly outperform naive heuristics in the double asymptotic limit where both problem size and constraints density tend to infinity. We revisit this picture by studying the finite-size behavior of some optimization algorithms across easy, intermediate, and hard regimes. Through rigorous analysis of large-graph asymptotics combined with numerical experiments on canonical problems (maximum independent set and maximum $K$-SAT), we demonstrate that while algorithms do eventually converge to theoretically predicted bounds, this convergence can be remarkably slow. In the intermediate regime where instances are already highly constrained, local algorithms achieve solutions substantially better than their predicted performance in the high-constraint-density limit. This gap between finite-regime and asymptotic behavior has important practical implications: sophisticated algorithmic design remains crucial even when asymptotic theory predicts inevitable~failure.
\end{abstract}

\section{Introduction}

Hard (combinatorial) optimization problems \cite{papadimitriou1998combinatorial} ask for the best value of an objective over a discrete, typically exponentially large set of feasible configurations. Satisfiability, routing and scheduling, graph partitioning, and covering/packing are canonical examples of this type of problems. They serve both as modeling primitives in operations research (e.g., logistics via routing-type formulations \cite{toth2014vehicle}) and as algorithmic backbones in automated reasoning and formal verification (e.g., SAT-based verification \cite{biere2009handbook}). Many such problems are NP-hard or NP-complete \cite{moore2011nature,garey2002computers}, and a large body of foundational work documents this hardness across optimization tasks.

A paradigmatic representative is the \emph{maximum independent set} (MIS) problem \cite{garey2002computers}. Given an undirected graph $G=(V,E)$, an independent set is a subset $I\subseteq V$ of pairwise non-adjacent nodes. An independent set is \emph{maximal} if no node can be added to it, and \emph{maximum} if its cardinality is the largest among all independent sets of $G$; this cardinality is the \emph{independence number} $\sigma(G)$. Computing $\sigma(G)$ is NP-hard \cite{karp2009reducibility}, yet the problem arises naturally in scheduling \cite{ambuhl2009single}, communication systems \cite{safar2007hard}, interference-free resource allocation in wireless networks \cite{sanghavi2009message}, or deposition dynamics in statistical physics \cite{jonckheere2021asymptotic}. Given an algorithm $\mathcal{A}$ that outputs an independent set of size $\sigma_\mathcal{A}(G)$, we quantify its quality via the \emph{performance ratio}
\begin{equation}
    \mathcal{P}_\mathcal{A}(G)=\frac{\sigma_\mathcal{A}(G)}{\sigma(G)}.\label{Perfration}
\end{equation}

A widely used approach to understanding the typical behavior of algorithms on such problems is average-case analysis on random instances drawn from well-defined graph ensembles \cite{frieze1997algorithmic}. For MIS one typically considers Erd\H{o}s-R\'enyi, random regular, or configuration-model graphs \cite{bollobas2011random} and studies the asymptotic independence ratio achieved by constructive procedures \cite{grimmett1975colouring}. On sparse Erd\H{o}s-R\'enyi graphs $G(n, \lambda/n)$ of mean degree $\lambda >0$, greedy heuristics that select nodes based on degree are provably near-optimal when $\lambda < e$ \cite{jonckheere2021asymptotic,wormald1995differential}, where $e$ denotes Euler's number. Analogous phase transitions have been established for configuration-model graphs with prescribed degree distributions \cite{jonckheere2021asymptotic}. When $\lambda > e$, degree-greedy optimality remains unproven, though it is widely conjectured, and supported by extensive numerical evidence, that greedy strategies become strictly sub-optimal.

Algorithms for MIS span a broad spectrum. At one end are sequential greedy heuristics \cite{frieze1990independence,grimmett1975colouring,johnson1973approximation}, which build an independent set by irrevocably adding nodes whose neighbors have not been selected; these admit efficient parallel implementations \cite{luby1985simple}. At the other end are message-passing methods such as belief propagation (BP) and its zero-temperature version, i.e. the max-product (MP) algorithm \cite{yedidia2003understanding}, a class of probabilistic algorithms rooted in statistical physics \cite{barbier2013hard,hartmann2006phase,mezard2009information,weigt2006message} which iteratively exchange local messages to approximate maximum a posteriori (MAP) assignments. Applied to weighted independent set formulations, MP can recover near-optimal configurations and, under certain conditions, solves an LP relaxation of the original problem, although convergence guarantees on loopy graphs remain limited \cite{wainwright2004tree,sanghavi2009message}. We will see in this paper that, when properly formulated, specific algorithms in these two classes can yield similar performance.

An influential perspective on algorithmic limits was articulated by Coja-Oghlan and Efthymiou \cite{coja2015independent}, who conjectured inherent barriers to surpassing simple heuristics on sparse random graphs based on intrinsic geometric properties of the space of solutions. For $G(n,\lambda/n)$ with average degree $\lambda \gg 1$, a trivial (degree-blind) greedy procedure finds an independent set of size $(1+o(1))\,n\ln(\lambda)/\lambda$ with high probability \cite{frieze1990independence,grimmett1975colouring}, which is asymptotically half the independence number (i.e. $\sigma(G)/2$). No polynomial-time algorithm is known to exceed this threshold by a factor $(1+\epsilon)$ for any fixed $\epsilon > 0$. The combinatorial geometry of the solution space in this regime exhibits features, notably the overlap gap property \cite{gamarnik2021overlap}, that appear to trap local and stable algorithms, leading to conjectures that no efficient algorithm can outperform the greedy baseline by a non-vanishing margin on such instances \cite{coja2015independent}. These conjectures parallel broader algorithmic-barrier hypotheses in random constraint satisfaction problems \cite{krzakala2007gibbs,mezard2002random}.

Despite these conjectured barriers, recent work \cite{wu2026unrealized} has reported evidence that the greedy benchmark can be exceeded on finite-size instances by simple polynomial-time procedures. In this paper, we investigate this phenomenon systematically for the MIS problem across several families of algorithms (sequential greedy heuristics and message-passing methods) and show that, while the asymptotic convergence to the half-optimality threshold does appear to hold, the approach to this limit is remarkably slow as a function of the mean degree $\lambda$. As a result, practical algorithmic performance can deviate significantly from asymptotic predictions even in regimes traditionally considered hard. To demonstrate that this slow-convergence phenomenon is not specific to graph problems, we complement our analysis with an analogous study of algorithms for the Max $K$-SAT problem \cite{mezard2002random}, highlighting the broader relevance of these conclusions across random combinatorial optimization. This conclusion could be particularly relevant due to the growing use of neural network-based solvers for combinatorial optimization, as in \cite{skenderi2026benchmarking,
angelini2023modern,wu2026unrealized}, whose empirical evaluations necessarily take place in the finite-size regime where classical baselines still perform well above their asymptotic limits.

The remainder of the paper is organized as follows. Section~\ref{sec:algorithms} introduces the algorithmic approaches studied: the max-product belief propagation algorithm for MIS, including its density-evolution description via population dynamics, and the sequential greedy algorithms (static degree greedy (SDG) and dynamic degree greedy (DG)). Section~\ref{sec:results} presents our main results for MIS: the asymptotic characterization of SDG performance (Theorem~\ref{thm:sdg}), finite-size simulations for DG, SDG, and MP on Erd\H{o}s-R\'enyi graphs, and a quantitative analysis of the rate at which the performance ratio converges to the half-optimality bound. Section~\ref{sec:ksat} extends the analysis beyond graph problems to the Max $K$-SAT setting, introducing the Random Clause (RC) algorithm, establishing its asymptotic performance (Theorem~\ref{thm:rc}), and presenting analogous finite-size results.

The code for all simulations can be found in the repository: \url{https://github.com/Alingoge/MIS-KSAT-code}.

\section{Algorithms for MIS and their asymptotic analysis}\label{sec:algorithms}

\subsection{Message passing: The max-product algorithm}

We employ max-product (MP), a message-passing algorithm for MAP estimation (see \cite{mezard2009information} for details), to find an MIS in the intermediate-hard regime.

Following \cite{sanghavi2009message}, the (weighted) MIS problem can be cast as MAP inference in a binary pairwise Markov random field. Given a graph $G=(V,E)$ with node weights $w_i>0$, associate to each node a binary variable $x_i\in\{0,1\}$, where $x_i=1$ indicates inclusion in the independent set and $x_i=0$ otherwise.
The joint distribution of a node assignment takes the form
\begin{equation}
    P(\mathbf{x})=\frac{1}{Z}\prod_{i\in V}\exp(w_i x_i)\prod_{(i,j)\in E}\mathbf{1}_{\{x_i x_j=0\}}.\label{postMIS}
\end{equation}
By construction, $P(\mathbf{x})$ assigns positive probability only to feasible independent set configurations, and its MAP maximizer therefore corresponds to a maximum-weight independent set.

\paragraph{Max-product (MP) algorithm.} At every iteration $t$, every node $i\in V$ sends a pair of messages $\{m_{i \rightarrow j}^{t}(0),m_{i \rightarrow j}^{t}(1)\}$ to each of its neighbors $j\in \partial i,$ where $\partial i$ represents the set of neighbors of $i$. Moreover, each node also maintains a set of beliefs $\{b_{i}^{t}(0),b_{i}^{t}(1)\}$, which serves as the core mechanism to determine whether or not a node $i$ should be included in the MIS according to its present state. To make implementation and analysis more straightforward, we perform the following transformation
\begin{equation}
    \gamma^t_{i\rightarrow j}=\ln\bigg( \frac{ m_{i \rightarrow j}^t(0)}{ m_{i \rightarrow j}^t(1)}\bigg).
\end{equation}
This reparameterization allows each node to receive only one message per neighbor and maintain a single belief. This modification of the max-product is often called the ``min-sum'' algorithm \cite{mezard2009information}, and is just a convenient reparameterization. In the rest of the paper, we refer to this as simply the max-product algorithm. The algorithm proceeds as follows:
\begin{enumerate}
    \item \textbf{Initialization.} Without loss of generality, we set $m_{i \rightarrow j}^{0}(0)=m_{i \rightarrow j}^{0}(1)=1$ at initialization. Therefore $\gamma_{i \rightarrow j}^{0} = 0$ for all $(i,j) \in E$.
    \item \textbf{Message updates.} For each directed edge $(i,j)$, update the message using the max-product recursion:
    \begin{equation}
    \label{eq:mis_minsum}
      \gamma_{i \rightarrow j}^{t+1} = \Bigl[ w_i - \sum_{u \in \partial i \setminus \{j\}} \gamma_{u \rightarrow i}^{t} \Bigr]_+ \quad \text{where} \quad [z]_+ := \max\{z,0\}.  
    \end{equation}
    \item \textbf{Log-belief ratio computation.} For each node $i$, compute the log-belief ratio:
    \begin{equation}
         \sigma_i^{t} = w_i - \sum_{u \in \partial i} \gamma_{u \rightarrow i}^{t}.
    \end{equation}
    \item \textbf{State assignment.} Based on the log-belief ratios, determine the new node states:
    \begin{equation}
      x_i^{t} = 
    \begin{cases}
        1 & \text{if } \sigma_i^{t} > 0,\\[4pt]
        0 & \text{if } \sigma_i^{t} < 0,\\[4pt]
        \frac{1}{2} & \text{if } \sigma_i^{t} = 0.
    \end{cases}   
    \end{equation}
    The value $x_i^{t}=1/2$ indicates an undecided state.
    \item \textbf{Convergence check.} Increment $t$ and repeat from step~2 until the configuration vector $\mathbf{x}^{t} = (x_i^{t})_{i \in V}$ stabilizes.
\end{enumerate}

When the algorithm is applied to a general loopy graph, it can lead to three possibilities: (i) the algorithm converges to the MAP estimate; (ii) the algorithm converges to a fixed point that does not correspond to the MAP assignment; or, (iii) the algorithm does not converge, which may occur in certain cases with too many cycles or other challenging structures in the graph. In practice, the algorithm often performs well and provides accurate approximate solutions for various graphical models and inference tasks. 

\paragraph{Asymptotic analysis of max-product by density evolution and population dynamics.}

On sparse Erd\H{o}s-R\'enyi graphs, neighborhoods are locally tree-like, enabling a \emph{density evolution} description of max-product in the large graph, $n\to\infty$, limit \cite{mezard2009information}. Let $\Gamma^t$ denote a random message on a typical (uniformly picked) directed edge at iteration $t$. For mean degree $\lambda$, the update induced by \eqref{eq:mis_minsum} takes the distributional form
\begin{equation}
\label{eq:de}
\Gamma^{t+1}\;\stackrel{\text{law}}{=}\;\Bigl[W-\sum_{r=1}^{D}\Gamma_r^{t}\Bigr]_+,
\end{equation}
where $D\sim\mathrm{Poisson}(\lambda)$, $(\Gamma_r^t)_r$ are independent and identically distributed (i.i.d.) copies of $\Gamma^t$, and $W$ is a random node weight drawn from $\text{Law}(w_i)$. Closed forms for \eqref{eq:de} are rarely available, so we approximate the evolving law of $\Gamma^t$ via \emph{population dynamics} (PD) \cite{mezard2009information}. The idea is to approximate the distribution of $\Gamma$ through a sample of $n$ i.i.d. copies of $\Gamma$. As $n \gg 1$, the empirical distribution of such a sample should converge to the actual distribution of $\Gamma$.
We present the PD implementation in the following algorithm \ref{population_dynamics_algorithm}.
\begin{algorithm}
\caption{Population Dynamics for Density Evolution}
\label{population_dynamics_algorithm}
\SetKwInOut{Input}{Input}
\SetKwInOut{Output}{Output}
\Input{
    $\lambda$: average degree, $n$: population size, $T$: maximum iterations, $f_w(x) = e^{-x}$:
}
\Output{
    $\Gamma^T$: final population of messages,
    $\sigma^T$: final population of beliefs.
}
\BlankLine
Initialize $\Gamma^0_i= 0$ and $\sigma^0_i =0$ for all $i\in [n]$.\\
\BlankLine
\For{$t = 1$ \KwTo $T$}{
    \ForEach{$i \in [1, n]$}{
        $k \gets \text{Sample from Poisson}(\lambda) + 1$ \\
        $w_i \gets f_w(k)$ \\
        Sample $k-1$ indices $\{i(1), i(2), \dots, i(k-1)\}$ independently and identically from $[n]$ \\
        $\Gamma_i^t \gets \big[w_i - \sum_{u=1}^{k-1} \Gamma_{i(u)}^{t-1}\big]_+$ \\
    }
    \ForEach{$i \in [1, n]$}{
        $d \gets \text{Sample from Poisson}(\lambda)$ \\
        $w_i \gets f_w(d)$ \\
        Sample $d$ indices $\{i(1), i(2), \dots, i(d)\}$ independently and identically from $[n]$ \\
        $\sigma_i^t \gets w_i - \sum_{u=1}^{d} \Gamma_{i(u)}^{t}$ \\
    }
}
\Return $\Gamma^T$, $\sigma^T$
\end{algorithm}

\subsection{Sequential greedy algorithms}

We consider \emph{sequential} algorithms that construct an independent set by iteratively exploring nodes and making irrevocable decisions. At each step $t\ge 0$, the node set is partitioned into three disjoint subsets
\begin{equation*}
V=\mathcal U_t \;\cup\; \mathcal A_t \;\cup\; \mathcal B_t,
\end{equation*}
where $\mathcal U_t$ are \emph{unexplored} nodes, $\mathcal A_t$ are \emph{active} nodes (selected into the independent set), and $\mathcal B_t$ are \emph{blocked} nodes (excluded because they are in the neighborhood of active nodes). The initialization is
\begin{equation*}
\mathcal U_0=V,\qquad \mathcal A_0=\emptyset,\quad \text{and} \quad \mathcal B_0=\emptyset.
\end{equation*}
Given a selection rule for choosing a node $i_t\in\mathcal U_{t-1}$, the update is
\begin{equation*}
\mathcal N(i_t)=\{j\in \mathcal U_{t-1}:\ (i_t,j)\in E\},
\end{equation*}
\begin{equation*}
    \mathcal A_t=\mathcal A_{t-1}\cup\{i_t\},\qquad  \mathcal B_t=\mathcal B_{t-1}\cup \mathcal N(i_t),\quad \text{and} \quad \mathcal U_t=\mathcal U_{t-1}\setminus\bigl(\{i_t\}\cup \mathcal N(i_t)\bigr).
\end{equation*}
The procedure stops at the first time $t^\star$ such that $\mathcal U_{t^\star}=\emptyset$. By construction, $\mathcal A_{t^\star}$ is a \emph{maximal} independent set.

\paragraph{Static degree greedy (SDG) algorithm.} 

Static degree greedy is a degree-based sequential algorithm. Let $d_i$ denote the degree of node $i$ in the \emph{original} graph $G$ (degrees are not updated during the run). At each iteration, SDG selects, uniformly at random, an unexplored node of minimum original degree. That is,
\begin{equation*}
i_t \in \arg\min_{i\in \mathcal U_{t-1}} d_i.
\end{equation*}
The sets $(\mathcal U_t,\mathcal A_t,\mathcal B_t)$ are then updated exactly as above using $\mathcal N(i_t)$. The algorithm terminates when $\mathcal U_t=\emptyset$, returning $\mathcal A_t$ as the output maximal independent set.

\paragraph{Dynamic degree greedy (DG) algorithm.}
Dynamic degree greedy is also a degree-based sequential algorithm but, in contrast to SDG, it selects nodes using the \emph{current} degree of nodes within the unexplored subgraph. Let $G[\mathcal U_{t-1}]$ denote the subgraph induced by the unexplored set at step $t-1$, and let
\begin{equation*}
d_i^{(t-1)} := \deg_{G[\mathcal U_{t-1}]}(i), \qquad i\in \mathcal U_{t-1},
\end{equation*}
be the degree of $i$ restricted to unexplored nodes. At iteration $t$, DG selects, uniformly at random, a node of $\mathcal U_{t-1}$ with minimum degree on $G[\mathcal U_{t-1}]$. That is,
\begin{equation*}
i_t \in \arg\min_{i\in \mathcal U_{t-1}} d_i^{(t-1)}.
\end{equation*}
It then updates $(\mathcal U_t,\mathcal A_t,\mathcal B_t)$ by activating $i_t$ and blocking its unexplored neighbors $\mathcal N(i_t)$ as before. DG terminates when $\mathcal U_t=\emptyset$, returning a maximal independent set $\mathcal A_t$. Compared to SDG, DG is more adaptive because node priorities evolve with the remaining graph structure.

    SDG differs from the DG rule only in the selection criterion: SDG uses fixed degrees from the original graph, whereas DG recomputes degrees on the induced subgraph of unexplored nodes. As every sequential algorithm, both run in at most $|V|$ iterations.

\paragraph{Asymptotic performance of SDG.} Let $G\sim \text{CM}_n(d_n)$ be a random graph drawn from the configuration model of size $n\geq1$ and degree sequence $d_n$, with the empirical distribution of $d_n$ converging weakly to some asymptotic degree distribution $(p_i)_{i\geq0}$. Here we will state our main theoretical result, which concerns the SDG algorithm. It allows us to track its performance in the large graph limit, which mirrors the density evolution analysis for max-product. For this, we will first define recursively an array of values that will appear in the theorem. For each $i,j\geq1$, define $m_i(j)$ according to
\begin{equation}\label{eq:recurs_1}
    \begin{cases}
        m_i(j) = 0 & \mbox{for } j \leq i \\
        m_i(j) = \frac{m_{i-1}(j)}{\left( 1 + \frac{m_{i-1}(i)}{u_{i-1}} i(i-2) \right)^{\frac{1}{i-2}}} & \mbox{for } j > i
    \end{cases};
\end{equation}
while, for phase-$2$, we get
\begin{equation}\label{eq:recurs_i}
    \begin{cases}
        m_2(j) = 0 & \mbox{for } j \leq 2 \\
        m_2(j) = m_{i-1}(j) e^{-2j \frac{m_1(2)}{u_1}} & \mbox{for } j > 2
    \end{cases}.
\end{equation}
Define a second auxiliary sequence according to
\begin{equation*}
    u_i = \begin{cases}
        \frac{u_{i-1}}{\left( 1 + \frac{m_{i-1}(i)}{u_{i-1}} i(i-2) \right)^{\frac{2}{i-2}}} & \mbox{for } i \neq 2 \\
        u_{i-1} e^{-2 \frac{m_1(2)}{u_1}} & \mbox{for } i=2
    \end{cases}.
\end{equation*}
The initial values of the sequences are fixed according to 
\begin{equation*}
    m_i(0) = p_i \quad (\text{for all } \,\, i=0,1,\dots)
\end{equation*}
and 
\begin{equation*}
    u_0 = \sum_{i=1}^\infty i p_i.
\end{equation*}
In terms of these sequences, we define, for all $i\geq1$,
\begin{equation}\label{eq:ind_set_phases}
    I_i = \begin{cases}
        p_0 & \mbox{for } i=0 \\
        \frac{u_{i-1}}{2i}\left( 1 - \left( 1 + \frac{m_{i-1}(i)}{u_{i-1}} i(i-2) \right)^{-\frac{2}{i-2}} \right) & \mbox{for } i\neq0,2 \\
        \frac{u_{1}}{4} \left( 1 - e^{-2\frac{m_1(2)}{u_1}} \right) & \mbox{for } i =2
    \end{cases}.
\end{equation}
We then have the following characterization of the asymptotic size of the independent sets found by the SDG algorithm.

\begin{theorem}\label{thm:sdg}
    Let $G\sim \text{CM}_n(d_n)$ be a random graph drawn from the configuration model of size $n\geq1$ and degree sequence $d_n$, with the empirical distribution of $d_n$ converging weakly to some asymptotic degree distribution $(p_i)_{i\geq0}$. Also assume that the second moment of the empirical distribution of $d_n$ converges to $\sum_{i\geq1} i^2 p_i$. If we denote the proportion of nodes contained in the independent set found by the static degree greedy algorithm run on $G$ by $\rho_{\text{SDG}}(G)$\footnote{where $\rho_{\text{SDG}}(G) = \sigma_{\text{SDG}}(G)/ n$.}, we then have that, as $n$ goes to infinity,
    \begin{equation*}
        \rho_{\text{SDG}}(G)\xrightarrow{\proba}\sum_{i=0}^\infty I_i.
    \end{equation*}
\end{theorem}
By means of this theorem, we can then compute the asymptotic fraction of nodes contained in independent sets constructed by SDG. We provide its proof in Appendix \ref{ap:proof_SDG} and use the result in the following section to study the asymptotic performance of SDG across different regimes of the mean degree

\section{Discussion of the results for MIS}\label{sec:results}

\begin{figure}[ht]
    \centering
    \includegraphics[width=0.6\linewidth]{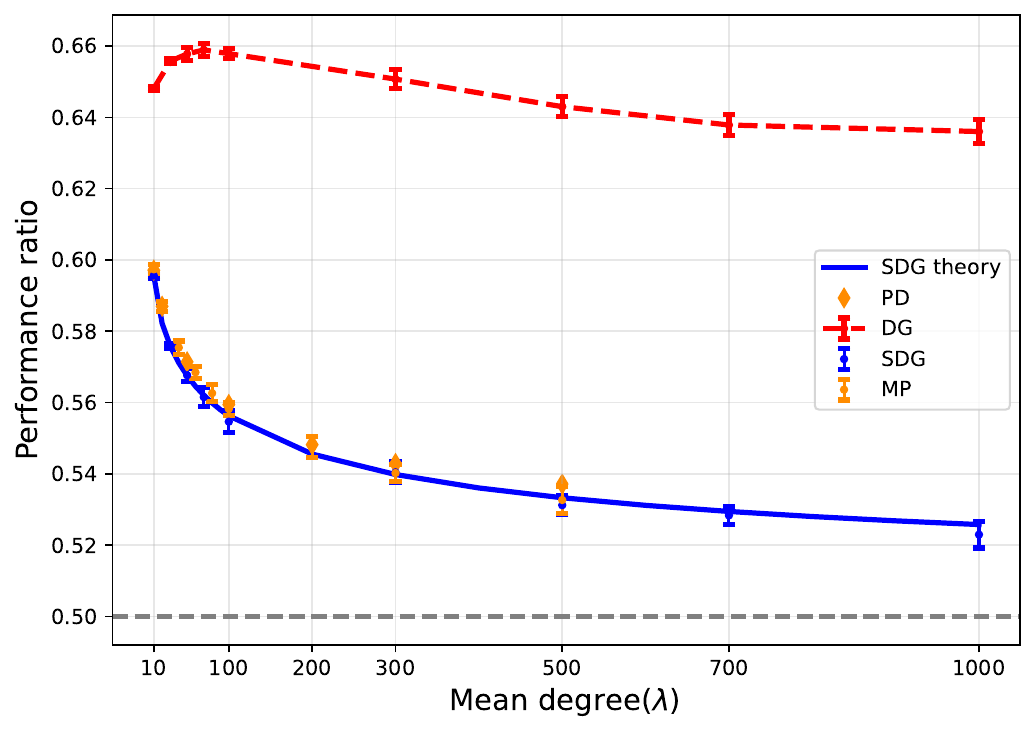}
        \caption{\textbf{Average performance ratio of the degree greedy (DG), static degree greedy (SDG), and max-product (MP) algorithms as a function of mean degree $\lambda$}. The solid blue line represents the theoretical asymptotic performance of SDG, while the blue dots with error bars represent its finite-size simulation results. The red dashed line represents the DG simulation results. The orange dots with error bars represent the MP numerical results, and the orange diamonds represent the population dynamics (PD) predictions. For all three algorithms, the finite-size performance remains above the 1/2 asymptotic threshold. This provides comprehensive (both theoretical and numerical) evidence that local algorithms can produce independent sets of size exceeding the theoretical threshold in sparse Erd\H{o}s-R\'enyi graphs even for practically large sizes and degree means. For DG and SDG, the results are averaged across $10$ realizations of Erd\H{o}s-R\'enyi graphs $G(n, \lambda/n)$ with $n = 10^5$, while the MP results are averaged over $25$ independent realizations of the same graph ensemble.}
    \label{fig:performance_ratio}
\end{figure}

We now present the results for the algorithms we introduced in Section~\ref{sec:algorithms} on a sparse Erd\H{o}s-R\'enyi graph $G(n,\lambda/n)$. Our primary objective is to compute the performance ratio of those algorithms. Interestingly, \cite{coja2015independent} shows that for $G(n,\lambda/n)$ with a mean degree of $\lambda \gg 1$, the independence number is approximately equal to $\sigma(G) \approx 2n\ln(\lambda)/\lambda$. Therefore, for sparse Erd\H{o}s-R\'enyi graphs of finite-but-large degree, the performance ratio $\mathcal{P}_{\mathcal{A}}$ will be approximated for our tested algorithms by replacing the denominator in \eqref{Perfration} by this approximation of $\sigma(G)$. 

\subsection{Results for sequential greedy algorithms}

We start with sequential algorithms run on instances of $G(n,\lambda/n)$. We consider values of $\lambda$ ranging from $10$ to $10^3$ and fix $n = 10^5$. On each graph instance, we apply the DG and SDG algorithms and record their performance ratios. For the SDG, we study both finite-size simulations and the asymptotic predictions provided by Theorem \ref{thm:sdg}, while for DG, we only study its finite-size simulations. Unlike SDG, which selects according to the original fixed degree sequence, DG selects by degree in the evolving subgraph $G[\mathcal U_{t-1}]$. Thus, analyzing DG asymptotically requires tracking the full empirical degree distribution of $G[\mathcal U_{t}]$, rather than a finite phase recursion, which we leave for future work. Figure~\ref{fig:performance_ratio} shows the average performance ratio of these algorithms over $10$ i.i.d. realizations of $G(n,\lambda/n)$.

We first discuss the DG results (red dots with error bars). The numerical simulations indicate that the DG algorithm produces independent sets whose sizes well exceed the theoretical threshold on average for sparse Erd\H{o}s-R\'enyi graphs. Recall that, according to existing theoretical results, no classical algorithm, either supported by rigorous analysis or numerical evidence, has been shown to find, with high probability in polynomial time, an independent set of size $(1+\epsilon)n\ln(\lambda)/\lambda$ for any fixed $\epsilon>0$. Similarly, no classical algorithm is known to achieve a performance ratio that exceeds $1/2$, represented by the gray horizontal dashed line in Figure~\ref{fig:performance_ratio}. Nevertheless, the results clearly show that incorporating degree information into the exploration process yields performance ratios above $1/2$. For all cases tested with $\lambda\gg 1$ and  $\lambda /n \ll 1$, the performance ratio approaches an asymptotic value of approximately $0.64$.

We next discuss the SDG results. First, the asymptotic predictions are in good agreement with the finite-size simulations, as shown by the close match between the blue solid line and the finite-size simulation points. Second, as expected, SDG performs worse than DG. Unlike DG, SDG is non-adaptive: it does not recompute node degrees on the induced subgraph of unexplored nodes. However, SDG also achieves performance ratios above the theoretical threshold $1/2$. Importantly, this improvement remains significant even for practically large values of $\lambda$. 

\begin{figure}[t]
\centering
\includegraphics[width=0.5\linewidth]{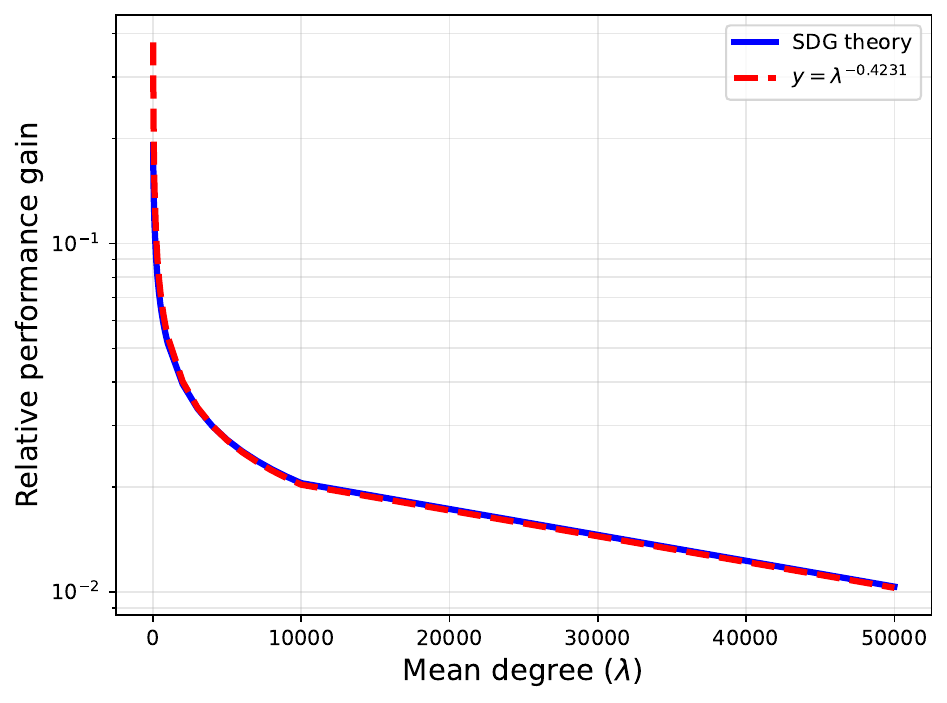}
\caption{\textbf{Convergence of the performance ratio of static degree greedy (SDG) towards the half-bound threshold}. The plot displays $\mathcal{R}_{\text{SDG}}(\lambda)$ given by \eqref{Rsdg}, interpreted as a natural performance lower bound for a large class of non-trivial algorithms for MIS, on Erd\H{o}s-R\'enyi graphs $G(n, \lambda/n)$ as a function of the degree $\lambda$ in the limit of large $n$. The blue solid line represents the convergence of the theoretical results of SDG, while the red dashed line represents a power law fit for large values of $\lambda$.}
\label{fig:linear_fit}
\end{figure}

SDG being arguably the simplest non-trivial algorithm for MIS, its asymptotic theoretical prediction provides a natural lower bound/baseline for more sophisticated algorithms, such as DG. This motivates quantifying the rate at which the asymptotic performance ratio of SDG converges to the theoretical threshold as a function of the mean degree $\lambda$. In Figure~\ref{fig:linear_fit}, we plot, on a log–linear scale, the quantity
\begin{equation}
\mathcal{R}_{\text{SDG}}(\lambda) =\frac{\sigma_{\text{SDG}}(\lambda)}{n \ln(\lambda) / \lambda} - 1, \label{Rsdg}
\end{equation}
that is, the relative performance gain above the half-threshold bound, as a function of $\lambda$.
To quantify the decay of this gain, we fit the data at large values of $\lambda$ \footnote{$\lambda \in [10^3, 5\times 10^4]$.} with a power-law function of the form $y = \lambda^{-a}.$ The fit yields a decay exponent $a \approx 0.4231$, indicating that the convergence of the performance gain toward zero is remarkably slow. Notably, there exists a broad range of large mean degrees over which the performance gain remains non‑negligible. This is even the case at $\lambda = 10^4$. This behavior is consistent with the intuition that, for a very large mean degree, the graph becomes increasingly homogeneous, with most nodes having approximately the same degree. In this regime, SDG should behave similarly to simple greedy dynamics which are degree-blind and thus expected to approach the best algorithmic bound rather slowly.

\begin{figure}[t]
\centering
\includegraphics[width=1\linewidth]{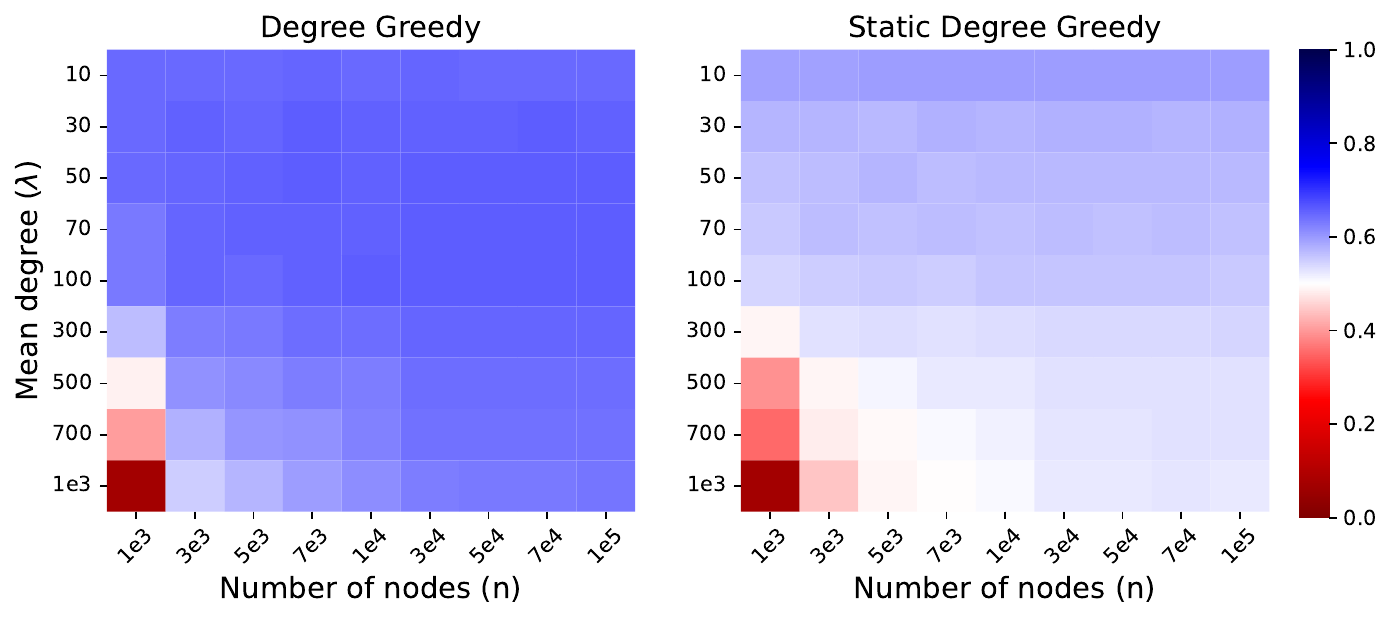}
\caption{\textbf{Heatmap of the performance ratios of the degree greedy (DG) and static degree greedy (SDG) algorithms on Erd\H{o}s-R\'enyi graphs $G(n, \lambda/n)$}. The left panel presents the results for DG, while the right panel shows the results for SDG. Both greedy algorithms achieve high ratios (above $1/2$) for sparser graphs, but the ratios fall below $1/2$ for denser graphs.}
\label{fig:heatmap}
\end{figure}

The performance gain observed for both DG and SDG is not restricted to the asymptotic regime $n \to \infty$. It remains nearly unchanged even for sparse finite-size graphs. Figure~\ref{fig:heatmap} illustrates this behavior through a heatmap of the performance ratio of the sequential algorithms for various values of $n$, ranging from $10^3$ to $10^5$. A similar performance gain is also observed for DG and other sophisticated classical and neural network-based algorithms in \cite{wu2026unrealized}.

\subsection{Results for max-product}

Before implementing MP, we introduced a modification that improves both its convergence and performance. A priori, MP is expected to belong to the class of algorithms that cannot generate, with high probability and in polynomial time, an independent set of size $(1+\epsilon)n\ln(\lambda)/\lambda$ in the limit of large $n$. Surprisingly, we find that parameterizing the local evidence with a positive weight that depends on the node's degree significantly improves MP's performance compared to the implementation of \cite{sanghavi2009message}. The weight is chosen as a monotonically decreasing function of a node's degree, so that nodes with fewer neighbors receive larger weights. We bias the algorithm toward selecting low-degree nodes by assigning to each node $i \in V$ a positive weight
\begin{equation}
\label{weight_function}
w_i = \exp(-|\partial i|).
\end{equation}
This modification encourages MP to construct independent sets consisting of weakly connected nodes, rather than treating all nodes as equally likely candidates. Such a prior also improves the stability of the message-passing iterations. However, this improvement does not hold for arbitrary monotonically decreasing functions of node degree. Empirically, we find that it is restricted to functions exhibiting at least exponential decay. In contrast, functions with polynomial decay do not induce a sufficiently strong bias and were observed to hinder the convergence of MP.  

In Figure~\ref{fig:performance_ratio}, we report the average performance ratio of MP over $25$ independent Erd\H{o}s-R\'enyi graphs, for $\lambda$ ranging from $10$ to $500$ and $n = 10^5$. The results are shown as orange points with error bars. We observe that the algorithm not only exceeds the theoretical threshold of $1/2$, but also closely matches the asymptotic behavior of SDG. To confirm the asymptotic behavior of MP, we employ a PD algorithm to approximate the message distribution in the large graph limit. The approximated message distribution is then used to estimate the density of nodes that satisfy the inclusion condition for the independent set. The orange diamonds represent the estimated performance ratio of the MP algorithm obtained from PD, using population sizes of $5 \times 10^6$ for all points except $\lambda=5\times 10^2$, where we used $10^7$ to obtain a better estimate.

Coming back to the close alignment between the asymptotic MP and SDG average performance on Erd\H{o}s-R\'enyi graphs, we find this result quite remarkable and interesting. Indeed, the two algorithms belong to fundamentally different families, with distinct implementation procedures and decision-making mechanisms. However, although MP and SDG both favor maximal independent sets containing low-degree nodes, the correspondence between the two is not exact. In SDG, this preference is explicit: at each step it selects uniformly among unexplored nodes of minimum original degree, so that the resulting maximal independent set can be seen as a random outcome of an exploration process that repeatedly exhausts the current minimum-degree class before moving to higher degrees. In contrast, MP does not impose an explicit exploration order, but the choice of weights introduces a strong energetic preference for including low-degree nodes. In fact, MP approximately searches for configurations $\mathbf{x}\in\{0,1\}^V$ maximizing $\sum_i w_i x_i$ under the hard constraints $\prod_{i,j}\mathbf{1}_{\{x_i x_j=0\}}$, and because of the choice of $w_i$ in \eqref{weight_function}, the objective becomes monotonic in the degree classes. More precisely, if one groups nodes by their original degree and writes $V=\bigcup_{d\ge 0}\mathcal V_d$, then the weighted objective can be decomposed as $\sum_{d\ge 0} e^{-d}\,|\{i\in \mathcal V_d: x_i=1\}|$. In the limit of infinitely fast decay (i.e., replacing $e^{-d}$ by $\varepsilon^d$ and letting $\varepsilon\downarrow 0$ or some other such limit), maximizing this quantity is equivalent to first maximizing the number of selected nodes in $\mathcal V_{d_{\min}}$, then, among those maximizers, maximizing the number selected in $\mathcal V_{d_{\min}+1}$, and so on, which amounts to a hierarchical optimization over conditional measures restricted to progressively higher degree classes. This is closely aligned with the SDG philosophy of saturating the minimum-degree class before proceeding to the next one. While MP operates through soft local messages rather than an explicit greedy schedule, the exponential weighting makes its effective target measure concentrate on independent sets that are ``as low-degree as possible''. This gives an intuition for why the two methods exhibit nearly identical performance in our experiments.

\section{Slow convergence beyond the MIS: the case of Max $K$-SAT}\label{sec:ksat}

The \emph{$K$-satisfiability} ($K$-SAT) problem consists of finding an assignment of $n$ Boolean variables $V_n = \{x_1, x_2, \ldots, x_n\}$ that satisfies a Boolean formula $\Phi_n$ in conjunctive normal form (CNF) comprising $m$ clauses. Each clause is a logical disjunction (OR) of exactly $K$ literals, where a literal is either a variable $x_i$ or its negation $\neg x_i$. In the random $K$-SAT problem, the $K$ literals in each clause are chosen uniformly at random, and the typical
hardness of the problem is controlled primarily by the clause-to-variable ratio $\alpha = m/n$, also called the clause density.

Standard SAT solvers offer limited information when a formula is unsatisfiable: they simply report that no solution exists, even though assignments that violate only a small number of clauses may be perfectly acceptable in practice. To address this limitation, one introduces a cost function that measures the fraction of satisfied clauses, and the goal becomes finding the assignment that maximizes it. This optimization variant, which remains well-defined even when no fully satisfying assignment exists, is known as \emph{Max $K$-SAT}~\cite{li2009maxsat}.

\subsection{Sequential algorithms for Max $K$-SAT}
The NP-hardness of Max $K$-SAT naturally leads to the use of sequential algorithms, which employ
greedy or randomized strategies to satisfy as large a fraction of clauses as possible. The simplest such algorithm is the \emph{random assignment}, in which each Boolean variable is independently set to true with probability $1/2$~\cite{vazirani2001approximation}; its performance serves as a baseline for any non-trivial algorithm. 

\paragraph{Random clause (RC) algorithm.}  Here, we analyze the behavior of a similar sequential algorithm which we call the random clause (RC) algorithm, and motivated by the fact that it follows a strategy analogous to that of SDG. It operates as follows. Consider $K \ge 2$ and $\alpha > 0$. The random $K$-CNF formula $\Phi_n = \Phi_n(K,\alpha)$ is a conjunction of $m = \lfloor \alpha n \rfloor$ clauses defined over the variable set $V_n = \{x_1, \dots, x_n\}$. Each clause is formed by selecting $K$ variables independently and uniformly at random from $V_n$, negating each independently with probability $1/2$. Starting from the empty assignment, the RC algorithm constructs a solution greedily as follows. At each step, it selects uniformly at random one undecided clause (i.e., a clause that is neither already satisfied nor falsified by the current partial assignment) then picks uniformly at random one of its currently unassigned literals and sets the corresponding variable so that the chosen literal evaluates to true. By construction, the selected clause is immediately satisfied; however, other undecided clauses that share the newly assigned variable may lose a free literal or become satisfied as a side effect. Note that the algorithm treats all undecided clauses uniformly, without any preference for clauses with fewer remaining free literals. After exactly $n$ steps, every variable has been assigned.

\paragraph{Asymptotic performance of RC.}

Here we state the main theoretical result regarding the RC algorithm for random Max $K$-SAT. In analogy with the SDG case, we first introduce the deterministic quantities that will appear in the theorem and that describe the limiting evolution of the algorithm.

Fix $K\ge 2$ and $\alpha>0$. Let $\Phi_n$ be a random $K$-CNF formula with $m=\lfloor \alpha n\rfloor$ clauses on $n$ variables, constructed with i.i.d. uniform variables and i.i.d. unbiased literal signs.

In order to describe the limiting performance of the algorithm, we introduce a family of deterministic functions $\big(u_j(t)\big)_{j=1,\dots,K}$, $s(t)$, and $f(t)$ where $t\in[0,1]$. Also define
\begin{equation*}
    U(t)=\sum_{j=1}^K u_j(t).
\end{equation*}
Define $x(t)=1-t$ to be the density of unassigned variables. These functions are defined as the unique solution, on the maximal interval where $x(t)>0$ and $U(t)>0$, of the following system of differential equations, with $j=1,\dots,K$,
\begin{equation}\label{eq:RC_system}
    \begin{dcases}
    \frac{d u_j}{dt} = - \frac{j u_j}{x} - \frac{u_j}{U} + \frac{(j+1)u_{j+1}}{2x} \\
    \frac{ds}{dt} = \sum_{j=1}^K\left(\frac{j u_j}{2x}+\frac{u_j}{U}\right) \\
    \frac{df}{dt} = \frac{u_1}{2x}
    \end{dcases}
\end{equation}

The initial conditions are determined by the structure of the random $K$-CNF at time $t=0$. Since initially all clauses have length $K$ and none are satisfied or falsified, we impose $u_K(0)=\alpha$, (for $j\leq K-1$) $u_j(0)=0$, $s(0)=0$, and $f(0) = 0$. In terms of this deterministic evolution, we may now characterize the asymptotic performance of the RC algorithm.

\begin{theorem}\label{thm:rc}
    Let $\Phi_n$ be a random $K$-CNF formula with $m=\lfloor \alpha n\rfloor$ clauses and let $\rho_{\mathrm{RC}}(\Phi_n)$ denote the proportion of clauses satisfied by the random clause algorithm applied to $\Phi_n$. If $t_* = \sup\{t\geq0: U(t) > 0\}$, then, as $n\to\infty$, we have
    \begin{equation*}
        \rho_{\mathrm{RC}}(\Phi_n) \xrightarrow{\proba} s(t_*),
    \end{equation*}
    with $U(t)$ and $s(t)$ as above.
\end{theorem}

Thus, the asymptotic satisfied-clause proportion achieved by the RC algorithm is completely characterized by the solution of the deterministic system \eqref{eq:RC_system}, in the same spirit as the SDG limit is characterized by the recursively defined quantities $(m_i(j),u_i,I_i)$. In Appendix \ref{app:proof_RC}, we present the proof of this theorem along with the dynamical interpretations of the equations in \eqref{eq:RC_system}. 

\subsection{Discussion of the results for RC}

We now present the results for the RC algorithm for random Max $K$-SAT. We consider clause densities $\alpha = m/n$ ranging over a logarithmically spaced grid, and we study both finite-size simulations and the asymptotic predictions provided by Theorem \ref{thm:rc}. All finite-size simulations correspond to instances with $n=10^3$ variables, and each data point is averaged over $10$ independent realizations.

Before discussing the simulations, we will argue that the performance of the random assignment of truth values is $1 - 2^{-K}$. If each variable is assigned independently and uniformly at random, a fixed $K$-clause is satisfied unless all its $K$ literals are false. Since the literals have independent unbiased signs and the assignment is uniform, the probability that a given literal is false is $1/2$, and therefore the probability that all $K$ literals are simultaneously false equals $2^{-K}$. Consequently, the expected fraction of satisfied clauses under random assignment is $1 - 2^{-K}$. By standard concentration arguments, in random $K$-SAT this fraction concentrates around $1 - 2^{-K}$ with high probability as $n\to\infty$. Hence $1 - 2^{-K}$ provides a natural baseline against which the performance of any algorithm should be compared. Similarly to the case of MIS, as the density of clauses $\alpha = m/n$ grows, we will see that the performance of RC slowly converges to that of a trivial solution: that is, to a random assignment.

\begin{figure}[t]
    \centering
    \includegraphics[width=0.49\linewidth]{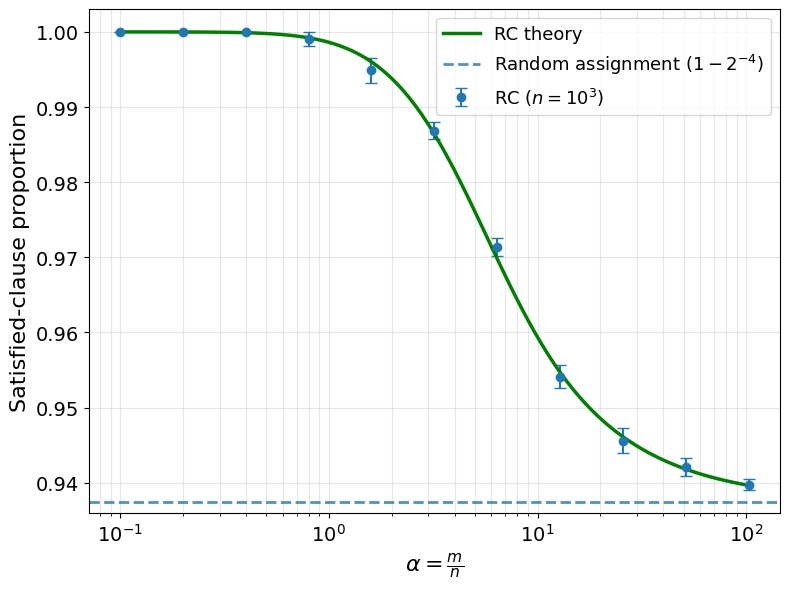} \hfill
    \includegraphics[width=0.49\linewidth]{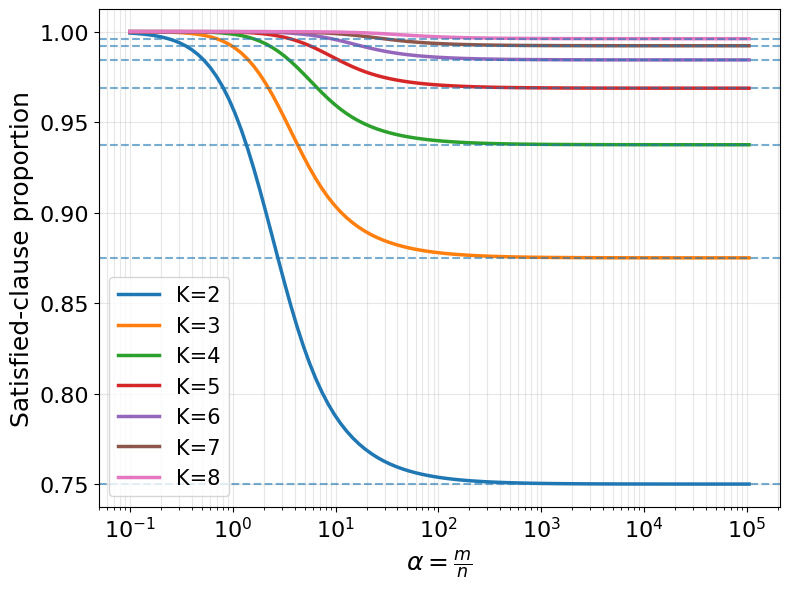}
    \caption{\textbf{Asymptotic performance of RC for Max $K$-SAT.} On the left, we compare the ratio of satisfied clauses as a function of the clause density ($\alpha=m/n$) of RC on simulations ($n=1000$, ten simulations for each point) with the predictions from the limit equations for $K=4$. Here we can appreciate that this ratio converges asymptotically, as argued, to the random assignment performance. On the right, asymptotic ratio of satisfied clauses as a function of clause density ($\alpha=m/n$) for different values of $K$. These simulations suggest that the convergence to the limit follows a similar rate of decay for all $K$ considered. In both graphs, the asymptotic performance of random assignments are plotted as dashed horizontal blue lines.}
\label{fig:simulationsRC}
\end{figure}

When the clause density $\alpha$ is large, each variable appears in a large number of clauses, and, by symmetry of the random model, approximately half of these appearances are as a positive literal and half as a negative literal. More precisely, for a fixed variable $x_i$, the numbers of clauses in which it appears positively and negatively are both of order $\alpha$, and their difference is negligible compared to their magnitude as $\alpha\to\infty$. As a consequence, assigning $x_i$ the value $1$ or $0$ produces, in expectation, nearly the same net effect on the total number of satisfied clauses. Since this balance becomes increasingly sharp at high density, any local advantage in choosing one value over the other vanishes asymptotically. The assignment decisions therefore become effectively neutral with respect to the global clause count. In this regime, each step of the RC dynamics behaves, from the perspective of the overall number of satisfied clauses, almost as if a value were chosen uniformly at random. This explains why, as $\alpha\to\infty$, the satisfied-clause proportion converges to the random assignment benchmark $1 - 2^{-K}$: in the high-density limit, random assignment is asymptotically optimal among strategies based only on such local information.

We now turn to the numerics. All of them can be found in Figure \ref{fig:simulationsRC}. In the first experiment, we fix $K=4$ and compare finite-size simulations with $n=10^3$ against the theoretical prediction obtained by solving the limiting ODE system of Theorem \ref{thm:rc}. For each value of $\alpha$, we compute the empirical mean and standard deviation of the satisfied-clause proportion achieved by RC and compare these with the asymptotic prediction. The results show very good agreement between theory and simulation, even at this moderate system size. The finite-size error bars remain small across the explored density range, and the theoretical curve closely tracks the empirical averages.

As $\alpha$ increases, both the simulations and the theoretical curve exhibit a gradual convergence toward the horizontal baseline $1 - 2^{-4} = 15/16$. This confirms the qualitative prediction discussed above: in the high-density regime, the RC algorithm becomes indistinguishable from random assignment in terms of its achieved satisfied-clause proportion.

In a second set of experiments, we compute the theoretical predictions of Theorem \ref{thm:rc} for several values of $K$, keeping $\alpha$ on a logarithmic scale. The resulting curves display the same qualitative behaviour for all $K$. For small and moderate clause densities, the RC algorithm significantly outperforms the random assignment baseline. However, as $\alpha$ grows, each curve slowly approaches its corresponding limit $1 - 2^{-K}$. The convergence toward this baseline is monotone and consistent across values of $K$, but it is slow on the logarithmic scale.

Even though the asymptotic theory predicts collapse to random assignment performance at high density, the approach to the limit occurs only for large values of $\alpha$. Over a wide intermediate range of clause densities, the RC algorithm maintains a non-trivial advantage over random assignment. In this sense, the situation parallels the MIS setting discussed previously: although a simple local algorithm eventually matches the trivial baseline in a dense regime, it retains a meaningful performance gap for practically relevant densities.

\section{Conclusion}
\label{sec:conclusion}

In this work, we have analyzed sequential greedy heuristics and message-passing algorithms for finding maximum independent sets in sparse Erdős-Rényi graphs. Through rigorous analysis of large-graph asymptotics, coupled with extensive numerical experiments, we have shown that for large graphs of large-but-finite degree, those algorithms typically find an independent set of size well above the theoretically predicted bound for local algorithms valid for diverging degree; this phenomenology was also discovered numerically by \cite{wu2026unrealized}. The theoretical threshold is ultimately attained, but the approach to this limit is remarkably slow as a function of the mean degree. As a result, for practical large mean degree (e.g. $\lambda = 10^4$), algorithmic performance can deviate significantly from asymptotic predictions even in regimes traditionally considered hard. We confirmed these conclusions for the Max  $K$-SAT problem \cite{mezard2002random}, highlighting the broader relevance of these conclusions across random combinatorial optimization. This gap between finite-regime and asymptotic behavior has important practical implications: sophisticated algorithmic design remains crucial, and evaluation practices need to be revisited accordingly.

Our conclusions are relevant within machine learning given the growing use of neural network-based solvers for combinatorial optimization \cite{skenderi2026benchmarking, angelini2023modern, wu2026unrealized}, whose empirical evaluations necessarily take place in the finite-but-large-size regime we study here. This offers a theoretical account of an already-observed empirical pattern: \cite{angelini2023modern} found that graph neural networks trained for maximum independent set are outperformed by simple degree-based greedy, and \cite{wu2026unrealized} found that leading AI-inspired methods are beaten not just by state-of-the-art classical solvers but sometimes by degree-greedy itself, with many non-backtracking neural methods implicitly reasoning like degree-greedy without matching its performance. Our results suggest an underlying reason: at finite size, degree-biased exploration exploits slow corrections to the asymptotic limit that a solver optimized only toward the asymptotic optimum will miss. Consequently, a neural network solver's distance from the asymptotic bound is a poor proxy for its practical quality. The relevant benchmark for a neural network solver is therefore not the asymptotic bound, but the finite-size performance already reachable by simple local algorithms — a standard that benchmarking efforts such as \cite{skenderi2026benchmarking} are well positioned to adopt.

\bibliographystyle{plain}
\bibliography{main}

\appendix

\section{Limit equations for the static degree greedy algorithm}
\label{ap:proof_SDG}

Let $G\sim\mbox{CM}_n(\bar{d})$ be a random graph drawn from the configuration model of size $n\geq1$ and degree sequence $\bar{d}$. Assume that $\mu(\cdot) := \sum_{i\leq n} \delta_{d_i}(\cdot)$ divided by $n$ converges weakly towards some asymptotic degree distribution $(p_k)_{k\geq1}$. For every $i\geq 1$ we will denote by $\mathcal{V}_i$ the set of nodes of degree $i$ in the graph $G$.

The SDG dynamics is divided into different stages that we will call \emph{phase-$1$}, \emph{phase-$2$}, etc. At first the dynamics is in phase-1 and only the unexplored nodes with degree $1$ in $G$ have an exponential clock. Thus, during this phase, only nodes of degree $1$ are added to the activated set. This phase goes on until the time $\tau_1 := \inf\{t\geq0: \mathcal{V}_1 \cap \mathcal{U}_t = \emptyset\}$ in which there are no more degree $1$ nodes to explore. After this time, phase-$2$ begins. Analogously, during phase-$2$, each unexplored degree-$2$ node has an exponential clock. When some of these clocks rings, the corresponding node and its neighborhood are explored in the way described above. This phase goes on until stopping time $\tau_2:=\inf\{t\geq\tau_1:\mathcal{V}_2\cap\mathcal{U}_t=\emptyset\}$. In general, if $i\geq1$ and we define inductively $\tau_i:=\inf\{t\geq\tau_{i-1}:\mathcal{V}_i\cap\mathcal{U}_t=\emptyset\}$, phase-$i$ starts after $\tau_{i-1}$ and finishes at time $\tau_i$ and during this time all the remaining unexplored nodes of degree $i$ are either activated or blocked.

For every $t\geq0$, we will let $\mu_t(\cdot):=\sum_{i\in\mathcal{U}_t}\delta_{d_i}(\cdot)$ be the degree counting measure of the unexplored nodes at time $t$ and $U_t:=\mu_t(\mathbb{N})+B_t$ be the total number of unpaired half-edges at time $t$ in $G$, where $\mu_t(\mathbb{N})= \sum_{j\geq 1} j\mu_t(j)$ is the number of unpaired half-edges of unexplored nodes at time $t$ and $B_t$ be the number of half-edges of blocked nodes that have not been paired at time~$t$.

To describe this dynamics as a Markov jump process in continuous time, we will use the \emph{principle of deferred decision} which implies that, no matter the order in which the half-edges that compose the graph are matched, if the matching is at each time done uniformly among the unmatched half-edges, the resulting graph has the same distribution. We will thus only match half-edges when required by the dynamics. Therefore, we will only reveal the graph when we need to do so. Using this, we will now show that the dynamics can be tracked according to a Markov jump process with respect to the quantities $(U_t,\mu_t)$. Suppose that, at a certain time $t\geq0$, the smallest degree of positive mass with respect to $\mu_t(\cdot)$ is $i\geq1$. We are then in the phase-$i$ of the exploration process. For simplifying the limiting equations, the rate of activation of each of the degree-$i$ nodes during phase-$i$ will be fixed (for every $t<\tau_i$) to be equal to $U_t/\mu_t(i)$.
    
Notice that when a clock rings we should match all of its half-edges and update the state of its unexplored neighbours. For each one of these matchings and $j\geq1$, there is a probability $j\mu_t(j)/U_t$ that the corresponding half-edge is matched to an unexplored node of degree $j$. Also, there is a probability $B_t/U_t = 1 - U^{-1}_t\sum_{i\geq1} i \mu_t(i)$ that it is matched to a half-edge of a blocked node. Moreover, if this half-edge is matched to a half-edge of a degree-$j$ node, the quantities of the process are modified in the following way
\begin{equation*}
    \begin{cases}
        U_t \longrightarrow U_t - 2 \\
        \mu_t(i) \longrightarrow \mu_t(i) - 1 \\
        \mu_t(j) \longrightarrow \mu_t(j) - 1;
    \end{cases}
\end{equation*}
while, if the half-edge of the activated node is matched to a half-edge of a blocked node, they are changed according to
\begin{equation*}
    \begin{cases}
        U_t \longrightarrow U_t - 2 \\
        \mu_t(i) \longrightarrow \mu_t(i) - 1.
    \end{cases}
\end{equation*}

By proofs similar to those in \cite{jonckheere2021asymptotic}, during the phase-$i$ the quantities $U_{t-\tau_{i-1}}/n$ and $\mu_{t-\tau_{i-1}}/n$ will converge in probability towards $u_t$ and $\bar{\mu}_t$ given by the solutions of
\begin{equation*}
    \begin{cases}
        \frac{d \bar{u}_t}{dt} = - 2 i \bar{u}_t  & \\
        \frac{d \bar{\mu}_t(i)}{dt} = -\left(\bar{u}_t + i^2 \bar{\mu}_t(i) \right) & \\
        \frac{d \bar{\mu}_t(j)}{dt} = - ij \bar{\mu}_t(j) & \mbox{for } j\neq i. \\
    \end{cases}
\end{equation*}
Let $\bar{t}_i := \inf\{t\geq0 : \bar{\mu}_t(i) = 0\}$. For each $i,j\geq1$, let $m_i(j)$ and $u_i$ be the number of nodes of initial degree $j$ which remain unexplored and of unexplored half-edges, respectively, at the end of phase $i$. These equations can then be easily integrated to obtain that, for $0 < t \leq \bar{t}_i$,
\begin{equation*}
    \begin{cases}
        \bar{u}_t = u_{i-1} e^{- 2 i t}  & \\
        \bar{\mu}_t(j) = 0 & \mbox{for } j < i \\
        \bar{\mu}_t(i) = -\frac{u_{i-1}}{i(i-2)} e^{-2i t} + \left( m_{i-1}(i) + \frac{u_{i-1}}{i(i-2)} \right) e^{-i^2 t} & \\
        \bar{\mu}_t(j) = m_{i-1}(j) e^{- i j t} & \mbox{for } j> i \\
    \end{cases}
\end{equation*}
for phase-$i$ (with $i\neq2$); and
\begin{equation*}
    \begin{cases}
        \bar{u}_t = u_1 e^{- 4 t}  & \\
        \bar{\mu}_t(j) = 0 & \mbox{for } j < 2 \\
        \bar{\mu}_t(2) = - u_1 t\, e^{-4t} + m_1(2)\, e^{-4t}& \\
        \bar{\mu}_t(j) = m_1(j) e^{- 2j t} & \mbox{for } j> 2 \\
    \end{cases}
\end{equation*}
for phase-$2$. From this we get that, for $i\neq2$
\begin{equation*}
    \bar{t}_i = \frac{1}{i(i - 2)}\, \ln\left( 1 + \frac{m_{i-1}(i)}{u_{i-1}} i(i-2) \right);
\end{equation*}
and
\begin{equation*}
    \bar{t}_2 = \frac{m_1(2)}{u_1}.
\end{equation*}

From the above explicit solutions, we get that, for $i\neq2$,
\begin{equation*}
    \begin{cases}
        m_i(j) = 0 & \mbox{for } j \leq i \\
        m_i(j) = \frac{m_{i-1}(j)}{\left( 1 + \frac{m_{i-1}(i)}{u_{i-1}} i(i-2) \right)^{\frac{1}{i-2}}} & \mbox{for } j > i
    \end{cases};
\end{equation*}
while, for phase-$2$, we get
\begin{equation*}
    \begin{cases}
        m_i(j) = 0 & \mbox{for } j \leq 2 \\
        m_2(j) = m_{i-1}(j) e^{-2j \frac{m_1(2)}{u_1}} & \mbox{for } j > 2
    \end{cases}.
\end{equation*}
In a similar way, we get a second recursion for the number of unexplored nodes by the end of phase-$i$ given by
\begin{equation*}
    u_i = \begin{cases}
        \frac{u_{i-1}}{\left( 1 + \frac{m_{i-1}(i)}{u_{i-1}} i(i-2) \right)^{\frac{2}{i-2}}} & \mbox{for } i \neq 2 \\
        u_{i-1} e^{-2 \frac{m_1(2)}{u_1}} & \mbox{for } i=2
    \end{cases}.
\end{equation*}
Through this recursion, the values of all the $m_i(j)$ can be inductively obtained.

This allows us to track the values of the asymptotic number of nodes of each degree through the evolution of the dynamics. But we are ultimately not interested in this but rather the size of the independent set obtained by the end of the exploration process. However, we are already in a position to derive an expression for this. Notice that during each phase, the number of nodes added to the independent set is equal to the number of half-edges matched divided by $2i$. This is true because during each exploration of a node, exactly $2i$ half-edges are matched. Thus, the limit of the number of nodes added to the independent set during phase-$i$ is asymptotically given by the difference in number of unmatched half-edges at time $\tau_{i-1}$ and $\tau_i$ divided by $2i$. Let $\mathcal{I}_i$ be the number of nodes added to the independent set during phase-$i$ divided by $n$. We then have that
\begin{equation*}
    \mathcal{I}_i = \begin{cases}
        p_0 & \mbox{for } i = 0 \\
        \frac{1}{2in}\left(U_{\tau_{i-1}}-U_{\tau_i}\right) & \mbox{for } i \geq 1
    \end{cases}.
\end{equation*}
By using that $\tau_i\xrightarrow{\proba}\sum_{j\leq i}\bar{t}_j$ and the limiting evolution of $U_t/n$ we then have that
\begin{equation*}
    \mathcal{I}_i \xrightarrow{\proba} \begin{cases}
        p_0 & \mbox{for } i=0 \\
        \frac{u_{i-1}}{2i}\left( 1 - \left( 1 + \frac{m_{i-1}(i)}{u_{i-1}} i(i-2) \right)^{-\frac{2}{i-2}} \right) & \mbox{for } i\neq0,2 \\
        \frac{u_{1}}{4} \left( 1 - e^{-2\frac{m_1(2)}{u_1}} \right) & \mbox{for } i =2
    \end{cases}.
\end{equation*}
This, then, proves the theorem.

\section{Limit equations for the random clause algorithm}\label{app:proof_RC}

The proof for the limit equations for Random Clause follows a similar strategy to that of SDG in the previous section but has certain characteristics that render it simpler. In particular, there is no need to embed the process in continuous time and choose an appropriate activation rate for the continuous time process. Furthermore, there are no phases in its dynamics meaning that the drift of the stochastic process remains the same throughout the exploration process. This, in particular, implies that there is no need for an explicit integration of the equations that would be later used to find limiting durations of each phase, as was the case with SDG.

Fix an integer $K\ge 2$ and a density parameter $\alpha>0$. For each $n\ge 1$, let $\Phi_n$ be a random $K$-CNF formula on the variable set $V_n=\{x_1,\dots,x_n\}$ with $m=\lfloor \alpha n\rfloor$ clauses. Each clause is generated independently by selecting $K$ variables uniformly from $V_n$ (with or without replacement, which does not affect the scaling limit) and assigning independent random signs to the corresponding literals, each sign being positive or negative with probability $1/2$. 

We consider the following sequential assignment dynamics, which we call the Random Clause (RC) algorithm. Starting from the empty assignment at step $\ell=0$, at each step $\ell=0,1,\dots,n-1$ we choose uniformly at random one clause among all clauses that are currently undecided, namely those clauses that are neither satisfied nor falsified by the current partial assignment. Inside the chosen clause we select uniformly one of its currently unassigned literals and assign its variable so that this literal becomes true. This rule is applied regardless of whether unit clauses are present: at every step the choice is uniform over all undecided clauses.

Let $\ell$ denote the discrete time index and set $t=\ell/n\in[0,1]$ for the associated rescaled time. A clause is said to have current length $j\in\{1,\dots,K\}$ at step $\ell$ if it is undecided and has exactly $j$ unassigned literals under the current partial assignment. We introduce the random variables $U^{(n)}_j(\ell)$ for the number of undecided clauses of current length $j$, $S^{(n)}(\ell)$ for the number of satisfied clauses, and $F^{(n)}(\ell)$ for the number of falsified clauses. We define the corresponding densities
\begin{equation*}
    u^{(n)}_j(t)=\frac{1}{n}U^{(n)}_j(\lfloor nt\rfloor), 
\end{equation*}
\begin{equation*}
    s^{(n)}(t)=\frac{1}{n}S^{(n)}(\lfloor nt\rfloor),
\end{equation*}
\begin{equation*}
    f^{(n)}(t)=\frac{1}{n}F^{(n)}(\lfloor nt\rfloor),
\end{equation*}
and the total density of undecided clauses
\begin{equation*}
    U^{(n)}(t)=\sum_{j=1}^K u^{(n)}_j(t).
\end{equation*}
At time $t=0$ we have $s^{(n)}(0)=f^{(n)}(0)=0$ and $u^{(n)}_K(0)\to \alpha$ in probability while $u^{(n)}_j(0)\to 0$ for $j\leq K-1$.

The process
\begin{equation*}
    \big(U^{(n)}_1(\ell),\dots,U^{(n)}_K(\ell),S^{(n)}(\ell),F^{(n)}(\ell)\big)
\end{equation*}
is Markovian. Indeed, once the current partial assignment and the current status of each clause are known, the distribution of the next step depends only on this information. As in the proof of the limit equations for SDG in the previous section, in order to compute the conditional expected increments of these variables, we invoke the principle of deferred decision. Conditional on the current filtration $\mathcal{F}_\ell$, the identities of the not-yet-exposed literals inside undecided clauses may be regarded as uniformly random among the $x(t)n$ unassigned variables, and the signs of those literals remain independent and unbiased. In particular, for a typical undecided clause of current length $j$, its $j$ unassigned variables behave, to leading order as $n\to\infty$, like $j$ independent uniform samples from the $x(t)n$ unassigned variables.

Fix $\ell$ and write $t=\ell/n$. We compute the conditional drift of $U^{(n)}_j(\ell)$. There are two distinct mechanisms that modify the number of undecided clauses of current length $j$ when the $(\ell+1)$-st assignment is performed.

First, the clause chosen by the RC rule is removed from the undecided pool and becomes satisfied. Since the choice is uniform among undecided clauses, the conditional probability that the chosen clause has current length $j$ equals
\begin{equation*}
    \frac{U^{(n)}_j(\ell)}{\sum_{r=1}^K U^{(n)}_r(\ell)}.
\end{equation*}
Whenever this occurs, $U^{(n)}_j$ decreases by one. This produces a contribution
\begin{equation*}
    -\frac{U^{(n)}_j(\ell)}{\sum_{r=1}^K U^{(n)}_r(\ell)}
\end{equation*}
to the conditional expectation of $\Delta U^{(n)}_j(\ell)=U^{(n)}_j(\ell+1)-U^{(n)}_j(\ell)$.

Second, we must account for all other clauses that contain the variable $v$ assigned at step $\ell+1$. Conditional on $\mathcal{F}_\ell$, let $v$ denote the variable set at this step. Consider a fixed undecided clause $C$ of current length $j$. Under deferred decision, the probability that $v$ appears among the $j$ unassigned literals of $C$ equals $j/(x n)+o(1/n)$. Summing over the $U^{(n)}_j(\ell)$ such clauses, the expected number of length-$j$ undecided clauses that contain $v$ is
\begin{equation*}
    \frac{j\,U^{(n)}_j(\ell)}{x n}+o(1).
\end{equation*}
Whenever $v$ appears in such a clause, that clause ceases to be length $j$ after the assignment. Indeed, since literal signs are independent and unbiased, the literal involving $v$ is satisfied with probability $1/2$ and falsified with probability $1/2$. In the first case the clause becomes satisfied and leaves the undecided pool; in the second case that literal is removed and the clause remains undecided but with current length $j-1$. In both cases the clause is no longer counted in $U^{(n)}_j$. Therefore this mechanism contributes an expected decrease
\begin{equation*}
    -\frac{j\,U^{(n)}_j(\ell)}{x n}+o(1)
\end{equation*}
to $\E[\Delta U^{(n)}_j(\ell)\mid\mathcal{F}_\ell]$.

Finally, clauses of current length $j+1$ may become length $j$ when they contain $v$ and the corresponding literal is falsified. The expected number of length-$(j+1)$ clauses containing $v$ is
\begin{equation*}
    \frac{(j+1)\,U^{(n)}_{j+1}(\ell)}{x n}+o(1).
\end{equation*}
Among these, half in expectation have the literal falsified, hence remain undecided and decrease their length by one. This yields an expected increase
\begin{equation*}
    \frac{(j+1)\,U^{(n)}_{j+1}(\ell)}{2x n}+o(1)
\end{equation*}
in $U^{(n)}_j$.

Combining the three contributions we obtain
\begin{equation*}
    \E\left[\Delta U^{(n)}_j(\ell)\mid\mathcal{F}_\ell\right] = -\frac{j\,U^{(n)}_j(\ell)}{x n} -\frac{U^{(n)}_j(\ell)}{\sum_{r=1}^K U^{(n)}_r(\ell)} + \frac{(j+1)\,U^{(n)}_{j+1}(\ell)}{2x n} + o(1),
\end{equation*}
with the convention $U^{(n)}_{K+1}\equiv 0$.

A similar reasoning gives the drift of $S^{(n)}$ and $F^{(n)}$. The satisfied count increases by one whenever the chosen clause is removed, and additionally by one for each clause containing $v$ whose literal is satisfied. The expected number of such additional satisfactions equals
\begin{equation*}
    \sum_{r=1}^K \frac{r\,U^{(n)}_r(\ell)}{2x n}+o(1).
\end{equation*}
Hence
\begin{equation*}
    \E\left[\Delta S^{(n)}(\ell)\mid\mathcal{F}_\ell\right] = 1 + \sum_{r=1}^K \frac{r\,U^{(n)}_r(\ell)}{2x n} + o(1).
\end{equation*}
A falsified clause can be created only when a unit clause contains $v$ and its unique literal is falsified. The expected number of unit clauses touched is $U^{(n)}_1(\ell)/(x n)+o(1)$, and half of them are falsified in expectation. Therefore
\begin{equation*}
    \E\left[\Delta F^{(n)}(\ell)\mid\mathcal{F}_\ell\right] = \frac{U^{(n)}_1(\ell)}{2x n}+o(1).
\end{equation*}

Dividing the drift equations by $n$ and passing formally to the limit suggests that the limiting deterministic functions $u_j(t)$, $s(t)$ and $f(t)$ satisfy, for $x(t)=1-t$ and $U(t)=\sum_{j=1}^K u_j(t)$,
\begin{equation*}
    \frac{du_j}{dt} = -\frac{j u_j}{x} -\frac{u_j}{U} +\frac{(j+1)u_{j+1}}{2x}, 
\end{equation*}
with $j=1,\dots,K$ and $u_{K+1}\equiv 0$, together with
\begin{equation*}
    \frac{ds}{dt} = \sum_{j=1}^K\left(\frac{j u_j}{2x}+\frac{u_j}{U}\right),
\end{equation*}
and
\begin{equation*}
    \frac{df}{dt} = \frac{u_1}{2x}.
\end{equation*}

To justify this convergence rigorously, observe that each assignment step changes the state by at most $O(1)$ clauses. Indeed, the number of clauses containing a fixed variable in a random $K$-CNF with $m=\alpha n$ clauses is asymptotically Poisson with mean $\alpha K$, hence is tight uniformly in $n$. Therefore the increments of each density coordinate are bounded by $O(1/n)$ with high probability. Furthermore, on any domain where $x(t)$ and $U(t)$ are bounded away from zero, the right-hand sides of the above ODE system are locally Lipschitz functions of $(u_1,\dots,u_K,s,f,t)$. 

By the standard differential equation method \cite{wormald1995differential}, it follows that for every $\varepsilon>0$ and for $t\in[\delta,1]$ on which $x(t)\ge\varepsilon$ and $U(t)\ge\varepsilon$, the process $(u^{(n)}_1(t),\dots,u^{(n)}_K(t),s^{(n)}(t),f^{(n)}(t))$ converges in probability, uniformly on $[0,T]$, to the unique solution of the above ODE system with initial conditions $u_K(0)=\alpha$, (for $J\leq K-1$) $u_j(0)=0$, $s(0)=0$, and $f(0)=0$. This finishes the proof of the theorem.

\end{document}